\documentclass[runningheads]{llncs}
\usepackage{graphicx}
\usepackage{amsmath,amssymb} 
\usepackage{color}
\usepackage[width=122mm,left=12mm,paperwidth=146mm,height=193mm,top=12mm,paperheight=217mm]{geometry}
\usepackage{multirow}
\usepackage{wrapfig}
\usepackage{url}
\begin{document}
\pagestyle{headings}
\mainmatter

\title{Deep Cascaded Bi-Network for \\ Face Hallucination\thanks{This paper is to appear in Proceedings of ECCV 2016.}} 

\titlerunning{Deep Cascaded Bi-Network for Face Hallucination}

\authorrunning{Shizhan Zhu \textit{et al.}}

\author{Shizhan Zhu$^1$, Sifei Liu$^{1,2}$, Chen Change Loy$^{1,3}$, Xiaoou Tang$^{1,3}$}


\institute{$^1$Department of Information Engineering, The Chinese University of Hong Kong \\ $^2$University of California, Merced \\ $^3$Shenzhen Institutes of Advanced Technology, Chinese Academy of Sciences}

\maketitle

\begin{abstract}
We present a novel framework for hallucinating faces of unconstrained poses and with very low resolution (face size as small as 5pxIOD\footnote{Throughout this paper, we use the inter-ocular distance measured in pixels (denoted as pxIOD), to concisely and unambiguously represent the face size.}).
In contrast to existing studies that mostly ignore or assume pre-aligned face spatial configuration (e.g. facial landmarks localization or dense correspondence field), we alternatingly optimize two complementary tasks, namely face hallucination and dense correspondence field estimation, in a unified framework.
In addition, we propose a new gated deep bi-network that contains two functionality-specialized branches to recover different levels of texture details.
Extensive experiments demonstrate that such formulation allows exceptional hallucination quality on in-the-wild low-res faces with significant pose and illumination variations.
\end{abstract}

\section{Introduction}
\label{intro}

Increasing attention is devoted to detection of small faces with an image resolution as low as 10 pixels of height~\cite{yang2015wider}.
Meanwhile, facial analysis techniques, such as face alignment~\cite{cao2014face,xiong2013supervised} and verification~\cite{schroff2015facenet,taigman2014deepface}, have seen rapid progress.
%
However, the performance of most existing techniques would degrade when given a low resolution facial image, because the input naturally carries less information, and images corrupted with down-sampling and blur would interfere the facial analysis procedure.
%
Face hallucination~\cite{jin2015robust,kolouri2015transport,yang2013structured,tappen2012bayesian,wang2005hallucinating,chakrabarti2007super,liu2007face,baker2000hallucinating}, a task that super-resolves facial images, provides a viable means for improving low-res face processing and analysis, e.g. person identification in surveillance videos and facial image enhancement.

\begin{figure}
\centering
\includegraphics[width=1\linewidth]{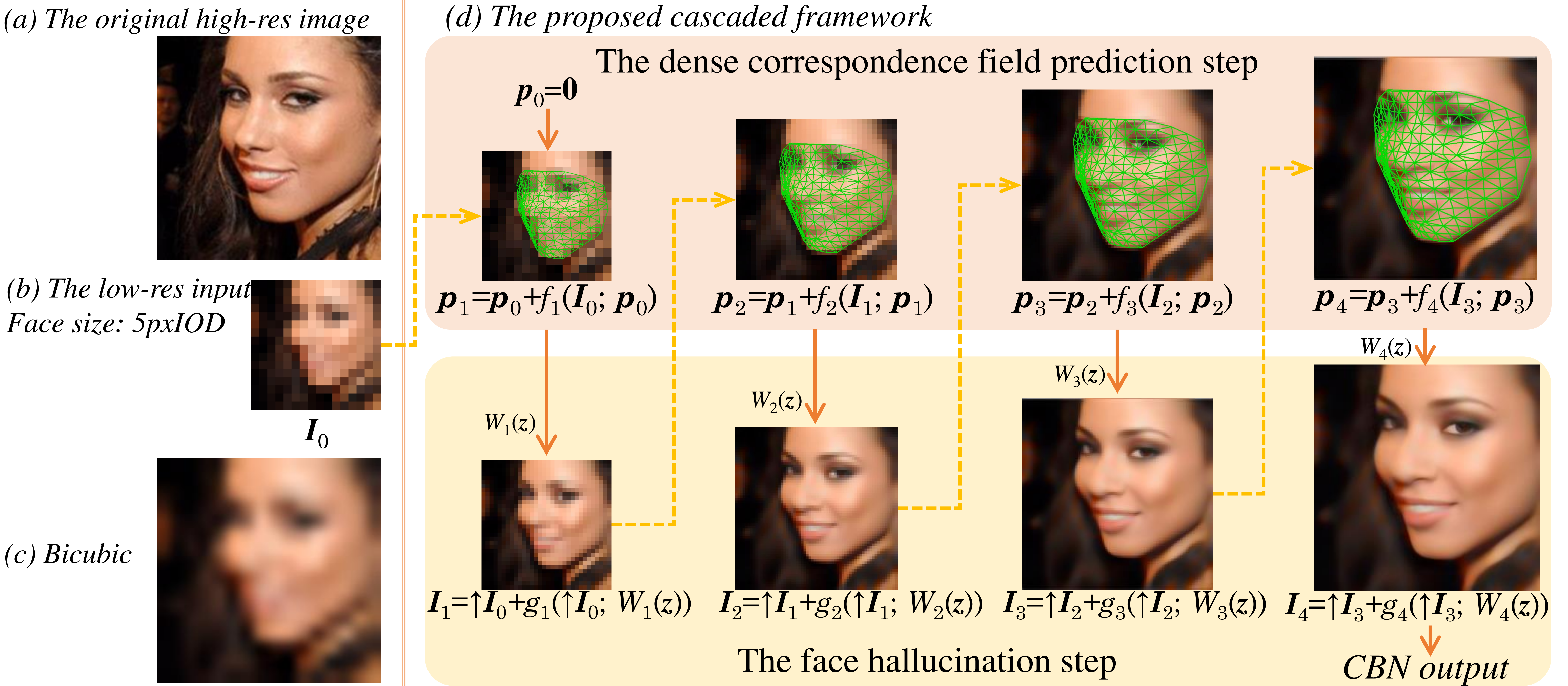}
\vskip -0.2cm
\caption{(a) The original high-res image. (b) The low-res input with a size of 5pxIOD. (c) The result of bicubic interpolation. (d) An overview of the proposed face hallucination framework. The solid arrows indicate the hallucination step that hallucinates the face with spatial cues, i.e. the dense correspondence field. The dashed arrows indicate the spatial prediction step that estimates the dense correspondence field.}
\label{fig:intro}
\vspace{-0.3cm}
\end{figure}

Prior on face structure, or face spatial configuration, is pivotal for face hallucination~\cite{liu2007face,tappen2012bayesian,jin2015robust}. %
The availability of such prior distinguishes the face hallucination task from the general image super-resolution problem~\cite{dong2015image,wang2015deep,huang2015single,gu2015convolutional,bruna2015super,salvador2015naive,dong2016accelerating,hui2016depth}, where the latter lacks of such global prior to facilitate the inference.
In this study, we extend the notion of prior to pixel-wise dense face correspondence field. We observe that an informative prior provides a strong semantic guidance that enables face hallucination even from a very low resolution. Here the dense correspondence field is necessary for describing the spatial configuration for its pixel-wise (not by facial landmarks) and correspondence (not by face parsing) properties. The importance of dense field will be reflected in Sec.~\ref{sr}.
An example is shown in Fig.~\ref{fig:intro} -- even an eye is only visible from a few pixels in a low-res image, one can still recover its qualitative details through inferring from the global face structure.

%


Nevertheless, obtaining an accurate high-res pixel-wise correspondence field is non-trivial given only the low-res input.
First, the definition of the high-res dense field is by itself ill-posed because the gray-scale of each pixel is distributed to adjacent pixels on the interpolated image (Fig.~\ref{fig:intro}(c)).
Second, the blur causes difficulties for many existing face alignment or parsing algorithms~\cite{xiong2013supervised,ren2014face,tzimiropoulos2015project,smith2013exemplar} because most of them rely on sharp edge information. 
Consequently, we face a chicken-and-egg problem - face hallucination is better guided by face spatial configuration, while the latter requires a high resolution face.
This issue, however, has been mostly ignored or bypassed in previous works (Sec.~\ref{related}).
%
%

In this study, we propose to address the aforementioned problem with a novel \textit{task-alternating cascaded framework}, as shown as Fig.~\ref{fig:intro}(d).
The two tasks at hand - the high-level face correspondence estimation and low-level face hallucination are complementary and can be alternatingly refined with the guidance from each other.
Specifically, motivated by the fact that both tasks are performed in a cascaded manner~\cite{cui2014deep,wang2015deep,tzimiropoulos2015project}, they can be naturally and seamlessly integrated into an alternating refinement process. 
During the cascade iteration, the dense correspondence field is progressively refined with the increasing face resolution, while the image resolution is adaptively upscaled guided by the finer dense correspondence field.
%

\begin{figure}[t]
\centering
\includegraphics[width=0.8\linewidth]{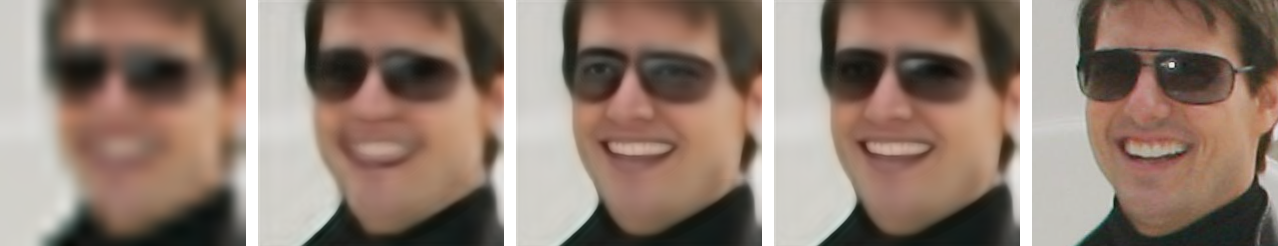}
\scriptsize
\medskip
(a)~Bicubic~~~~~~(b)~Common~~~~(c)~High-Freq.~~~~~(d)~\textbf{CBN}~~~~~~~(e)~Original
\medskip
\vspace{-0.7cm}
\caption{Examples for visualizing the effects of the proposed gated deep bi-network. (a) The bicubic interpolation of the input. (b) Results where only common branches are enabled. (c) Results where only high-frequency branches are enabled. (d) Results of the proposed CBN when both branches are enabled. (e) The original high-res image. Best viewed by zooming in the electronic version.}
\label{fig:wo}
\vspace{-0.3cm}
\end{figure}

To better recover different levels of texture details on faces, we propose a new \textit{gated deep bi-network} architecture in the face hallucination step in each cascade.
Deep convolutional neural networks have demonstrated state-of-the-art results for image super resolution~\cite{dong2015image,wang2015deep,gu2015convolutional,bruna2015super}.
In contrast to aforementioned studies, the proposed network consists two functionality-specialized branches, which are trained end-to-end. The first branch, referred as common branch, conservatively recovers texture details that are only detectable from the low-res input, similar to general super resolution.
The other branch, referred as high-frequency branch, super-resolves faces with the additional high-frequency prior warped by the estimated face correspondence field in the current cascade. Thanks to the guidance of prior, this branch is capable of recovering and synthesizing un-revealed texture details in the overly low-res input image.
A pixel-wise gate network is learned to fuse the results from the two branches.
Figure~\ref{fig:wo} demonstrates the properties of the gated deep bi-network.
As can be observed, the two branches are complementary. Although the high-frequency branch synthesizes the facial parts that are occluded (the eyes with sun-glasses), the gate network automatically favours the results from the common branch during fusion.

We refer the proposed framework as \textit{Cascaded Bi-Networks} (CBN) hereafter.
%
%
%
We summarize our contribution as follows:
%
\begin{enumerate}
  \item  While conducting face hallucination or dense face correspondence field is hard on low-res images, we circumvent this problem through a novel task-alternating cascade framework. In comparison to existing approaches, this framework has an appealing property of not assuming pre-aligned inputs or any availability of spatial information (e.g. landmark, parsing map).
  \item We propose a gated deep bi-network that can effectively exploit face spatial prior to recover and synthesize texture details that even are not explicitly presented in the low-resolution input.
  \item We explore the lower bound of the input face resolution for recovering reasonable and high quality details, and provide extensive results and discussion.
\end{enumerate}

We perform extensive experiments against general super-resolution and face hallucination approaches on various benchmarks. Our method not only achieves high Peak Signal to Noise Ratio (PSNR), but also superior quality perceptually. 
Demo codes will be available in our project page \url{http://mmlab.ie.cuhk.edu.hk/projects/CBN.html}.


\vspace{-0.2cm}

\section{Related Work}
\label{related}

\vspace{-0.1cm}

%
%
%
%

\noindent \textbf{Face hallucination and spatial cues}. There is a rich literature in face hallucination~\cite{jin2015robust,kolouri2015transport,yang2013structured,tappen2012bayesian,wang2005hallucinating,chakrabarti2007super,liu2007face,baker2000hallucinating}.
Spatial cues are proven essential in most of previous works, and are utilized in various forms.
For example, Liu et al.~\cite{liu2007face,tappen2012bayesian} and Jin et al.~\cite{jin2015robust} devised a warping function to connect the face local reconstruction with the high-res faces in the training set. However, a low-res correspondence field\footnote{We assume that we only correspond from pixel to pixel.} may not be sufficient for aiding the high-res face reconstruction process, while obtaining the high-res correspondence field is ill-posed with only a low-res face given.
Yang et al.~\cite{yang2013structured} assumed that facial landmarks can be accurately estimated from the low-res face image. This is not correct if the low-res face is rather small (e.g. 5pxIOD), since the gray-scale is severely distributed to the adjacent pixels (Fig.~\ref{fig:intro}(c)).
Wang et al.~\cite{wang2005hallucinating} and Kolouri et al.~\cite{kolouri2015transport} only aligned the input low-res faces with an identical similarity transform (e.g. the same scaling and rotation). Hence these approaches can only handle canonical-view low-res faces.
Zhou et al.~\cite{zhou2015learning} pointed out the difficulties of predicting the spatial configuration over a low-res input, and did not take any spatial cues into account for hallucination.
In contrast to all aforementioned approaches, we adaptively and alternatingly estimate the dense correspondence field as well as hallucinate the faces in a cascaded framework. The two mutual tasks aid each other and hence our estimation of the spatial cues and hallucination can be better refined with each other.

\noindent \textbf{Cascaded prediction} The cascaded framework is privileged both for image super-resolution (SR)~\cite{wang2015deep,cui2014deep} and facial landmark detection~\cite{dollar2010cascaded,cao2014face,xiong2013supervised,ren2014face,zhang2014coarse,tzimiropoulos2015project,zhu2015face,Zhu_2016_CVPR}.
For image SR, Wang et al.~\cite{wang2015deep} showed that two rounds of $2\times$ upscaling is better than a single round of $4\times$ upscaling in their framework.
For facial landmark detection, the cascaded regression framework has revolutionized the accuracy and has been extended to other areas~\cite{wang2015tric}.
The key success of the cascaded regression comes from its coarse-to-fine nature of the residual prediction. As pointed out by Zhang et al.~\cite{zhang2014coarse}, the coarse-to-fine nature can be better achieved by the increasing facial resolution among the cascades.
To our knowledge, no existing work has integrated these two closely related tasks into a unified framework.

\noindent \textbf{The bi-network architecture} The bi-network architecture~\cite{tenenbaum2000separating,pirsiavash2009bilinear,xiong2015recognize} has been explored in various form, such as bilinear networks~\cite{lin2015bilinear,gao2016compact} and two-stream convolutional network~\cite{simonyan2014two}.
In \cite{lin2015bilinear}, the two factors, namely object identification and localization, are modeled by the two branches respectively.
This is different from our model, where the two factors, the low-res face and the prior, are jointly modeled in one branch (the high-frequency branch), while the other branch (the common branch) only models the low-res face.
In addition, the two branches are joined via the gate network in our model, different from the outer-production in \cite{lin2015bilinear}.
In \cite{simonyan2014two}, both spatial and temporal information are modeled by the network, which is different from our model, where no temporal information is incorporated.
Our architecture also differs from \cite{zhou2015learning}. In \cite{zhou2015learning}, the output is the average weighted by a scalar between the result of one branch and the low-res input. Moreover, neither of the two branches utilizes any spatial cues or prior in \cite{zhou2015learning}.

\vspace{-0.2cm}

\section{Cascaded Bi-Network (CBN)}
\label{method}

\vspace{-0.1cm}

\subsection{Overview}
\label{overview}
\vspace{-0.1cm}


\noindent \textbf{Problem and notation}.
Given a low-resolution input facial image, our goal is to predict its high-resolution image.
%
%
We introduce the two main entities involved in our framework:

The \textit{facial image} is denoted as a matrix $\mathbf{I}$. We use $\mathbf{x} \in \mathbb{R}^2$ to denote the $(x,y)$ coordinates of a pixel on $\mathbf{I}$.

\begin{figure}[t]
\centering
\includegraphics[height=3cm]{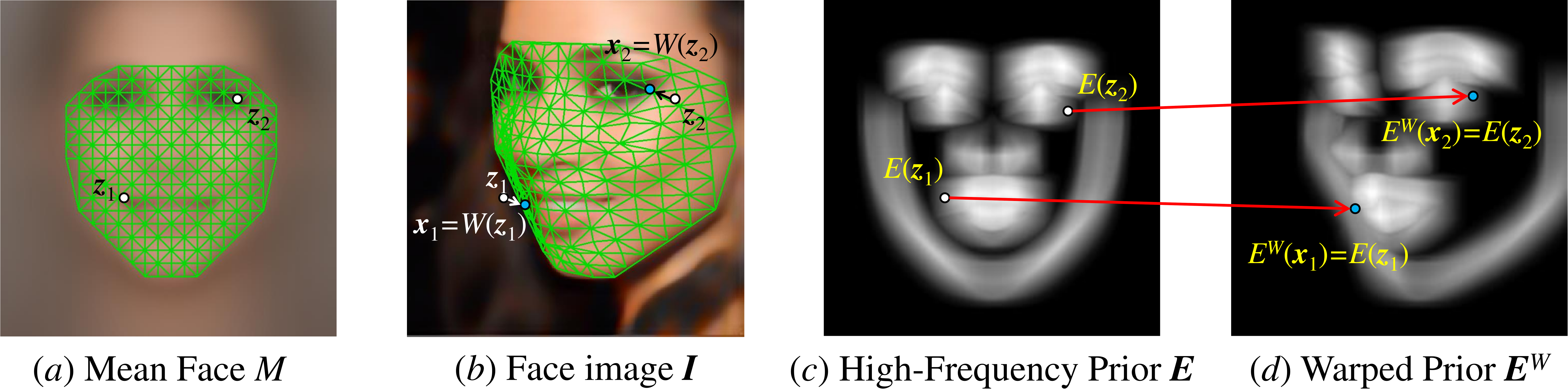}
\vspace{-0.3cm}
\caption{(a,b) Illustration of the mean face template $M$ and the facial image $\mathbf{I}$. The grid denotes the dense correspondence field $W(\mathbf{z})$. The warping from $\mathbf{z}$ to $\mathbf{x}$ is determined by this warping function $W(\mathbf{z})$. (c,d) Illustration of the high-frequency prior $\mathbf{E}$ and the prior after warping  $\mathbf{E}^W$ for the sample image in (b). Note that both $\mathbf{E}$ and $\mathbf{E}^W$ have $C$ channels. Each channel only contains one `contour line'. For the purpose of visualization, in this figure, we reduce their channel dimension to one channel with $\max$ operation. We leave out all indices $k$ for clarity. Best viewed in the electronic version.}
\label{fig:prior}
\end{figure}

The \textit{dense face correspondence field} defines a pixel-wise correspondence mapping from $M \subset \mathbb{R}^2$ (the 2D face region in the mean face template) to the face region in image $\mathbf{I}$.
We represent the dense field with a warping function~\cite{baker2004lucas}, $\mathbf{x} = W(\mathbf{z}): M \rightarrow \mathbb{R}^2$, which maps the coordinates $\mathbf{z} \in M$ from the mean shape template domain to the target coordinates $\mathbf{x} \in \mathbb{R}^2$.
See Fig.~\ref{fig:prior}(a,b) for a clear illustration.
Following \cite{snape2015face}, we model the warping residual $W(\mathbf{z}) - \mathbf{z}$ as a linear combination of the dense facial deformation bases, i.e.
\begin{align}
W(\mathbf{z}) = \mathbf{z} + \mathbf{B}(\mathbf{z}) \mathbf{p}
\label{eqn:w}
\end{align}
where $\mathbf{p} = [p_1 \dots p_N]^\top \in \mathbb{R}^{N \times 1}$ denotes the deformation coefficients and $\mathbf{B}(\mathbf{z}) = [\mathbf{b}_1(\mathbf{z}) \dots \mathbf{b}_N(\mathbf{z})] \in \mathbb{R}^{2 \times N}$ denotes the deformation bases.
The $N$ bases are chosen in the AAMs manner~\cite{cootes2001active}, that 4 out of $N$ correspond to the similarity transform and the remaining for non-rigid deformations.
Note that the bases are pre-defined and shared by all samples. Hence the dense field is actually controlled by the deformation coefficients $\mathbf{p}$ for each sample.
When $\mathbf{p} = \mathbf{0}$, the dense field equals to the mean face template.

We use the hat notation ($~\hat{ }~$) to represent ground-truth in the learning step. For example, we denote the high-resolution training image as $\hat{\mathbf{I}}$.

\vspace{0.15cm}
\noindent \textbf{Framework overview}.
We propose a principled framework to alternatively refine the face resolution and the dense correspondence field.
Our framework consists of $K$ iterations (Fig.~\ref{fig:intro}(d)). Each iteration updates the prediction via
\begin{align}
&\mathbf{p}_k = \mathbf{p}_{k-1} + f_k(\mathbf{I}_{k-1}; ~\mathbf{p}_{k-1}); W_k(\mathbf{z}) = \mathbf{z} + \mathbf{B}_k(\mathbf{z}) \mathbf{p}_k; \label{eqn:p_overview}\\
&\mathbf{I}_k = \hspace{0.1cm} \uparrow \hspace{-0.1cm} \mathbf{I}_{k-1} + g_k(\uparrow \hspace{-0.1cm} \mathbf{I}_{k-1}; ~W_k(\mathbf{z})); ~~~~~~~~~~~ (\forall \mathbf{z} \in M_k), \label{eqn:I_overview}
\end{align}
where $k$ iterates from 1 to $K$. Here, Eq.~\ref{eqn:p_overview} represents the dense field updating step while Eq.~\ref{eqn:I_overview} stands for the spatially guided face hallucination step in each cascade.
`$\uparrow$' denotes the upscaling process ($2\times$ upscaling with bicubic interpolation in our implementation).
All the notations are now appended with the index $k$ to indicate the iteration.
A larger $k$ in the notation of $\mathbf{I}_k$, $W_k$, $\mathbf{B}_k$ and $M_k$\footnote{We also append the subscript $k$ for $M$ because the mean face template domain $M_k$ do not have the same size in different iteration $k$.} indicates the larger resolution and the same $k$ indicates the same resolution.
The framework starts from $\mathbf{I}_0$ and $\mathbf{p}_0$.
$\mathbf{I}_0$ denotes the input low-res facial image.
$\mathbf{p}_0$ is a zero vector representing the deformation coefficients of the mean face template.
The final hallucinated facial image output is $\mathbf{I}_K$.

\begin{figure}[t]
\centering
\includegraphics[width=\linewidth]{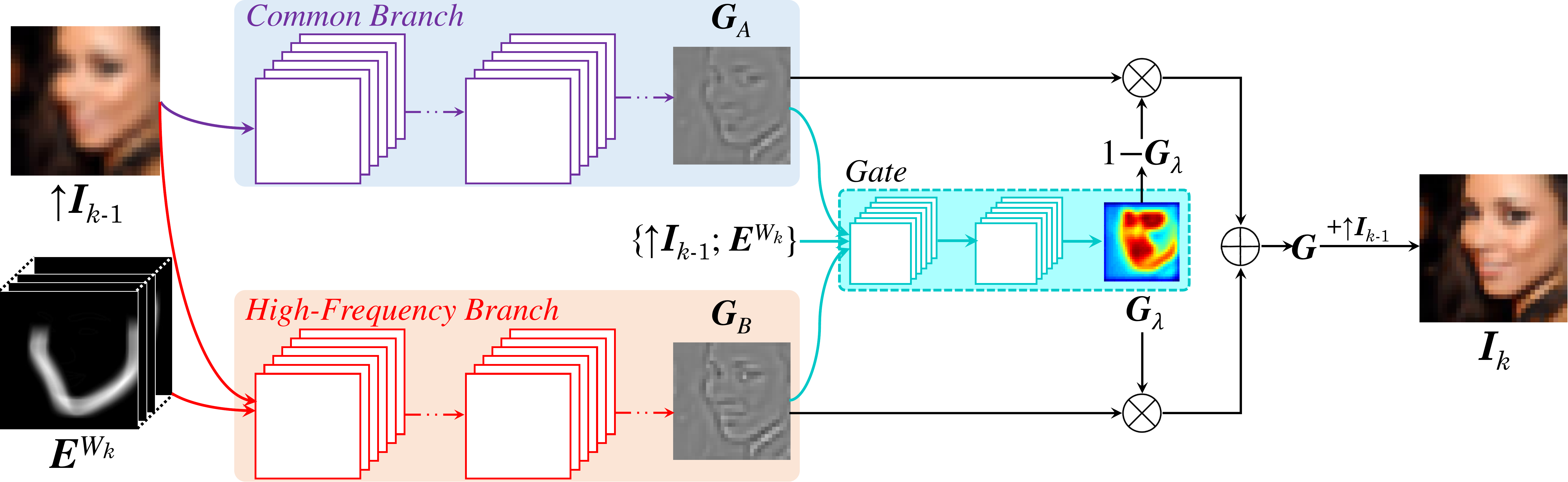}
\vspace{-0.8cm}
\caption{Architecture of the proposed deep bi-network (for the $k$-th cascade). It consists of a common branch (blue), a high-frequency branch (red) and the gate network (cyan).}
\vspace{-0.35cm}
\label{fig:network}
\end{figure}

\vspace{0.15cm}
\noindent \textbf{Model, inference and learning}.
Our model is composed of functions $f_k$ (dense field estimation) and $g_k$ (face hallucination with spatial cues).
The deformation bases $\mathbf{B}_k$ are pre-defined for each cascade and fixed during the whole training and testing procedures.
During testing, we repeatedly update the image $\mathbf{I}_k$ and the dense correspondence field $W_k(\mathbf{z})$ (basically the coefficients $\mathbf{p}_k$) with Eq.~\ref{eqn:p_overview}, \ref{eqn:I_overview}.
The learning procedure works similarly to the inference but incorporating the learning process of the two functions - $g_k$ for hallucination and $f_k$ for predicting the dense field coefficients. We present their learning procedures in Sec.~\ref{sr} and Sec.~\ref{flow} respectively.
%

\vspace{-0.25cm}
\subsection{$g_k$ - Gated deep bi-network: Face hallucination with spatial cues}
\label{sr}

We propose a gated deep bi-network architecture for face hallucination with the guidance from spatial cues.
We train one gated bi-network for each cascade.
For the $k$-th iteration, we take in the input image $\uparrow \hspace{-0.1cm} \mathbf{I}_{k-1}$ and the current estimated dense correspondence field $W_k(\mathbf{z})$, to predict the image residual $\mathbf{G} = \mathbf{I}_k-\uparrow~\hspace{-0.22cm}\mathbf{I}_{k-1}$.
%


As the name indicates, our gated bi-network contains two branches.
%
%
In contrast to \cite{lin2015bilinear} where two branches are joined with outer production, we combine the two branches with a gate network.
More precisely, if we denote the output from the common branch (A) and the high-frequency branch (B) as $\mathbf{G}_A$ and $\mathbf{G}_B$ respectively, we combine them with
\begin{align}
g_k(\uparrow \hspace{-0.1cm} \mathbf{I}_{k-1}; ~W_k(\mathbf{z})) = \mathbf{G} = (\mathbf{1} - \mathbf{G}_\lambda) \otimes \mathbf{G}_A + \mathbf{G}_\lambda \otimes \mathbf{G}_B,
\end{align}
where $\mathbf{G}$ denotes our predicted image residual $\mathbf{I}_k - \uparrow \hspace{-0.1cm} \mathbf{I}_{k-1}$ (i.e. the result of $g_k$), and $\mathbf{G}_\lambda$ denotes the pixel-wise soft gate map that controls the combination of the two outputs $\mathbf{G}_A$ and $\mathbf{G}_B$. We use $\otimes$ to denote element-wise multiplication.

Figure~\ref{fig:network} provides an overview of the gated bi-network architecture.
Three convolutional sub-networks are designed to predict $\mathbf{G}_A$, $\mathbf{G}_B$ and $\mathbf{G}_\lambda$ respectively.
The common branch sub-network (blue in Fig.~\ref{fig:network}) takes in only the interpolated low-res image $\uparrow \hspace{-0.1cm} \mathbf{I}_{k-1}$ to predict  $\mathbf{G}_A$ while the high-frequency branch sub-network (red in Fig.~\ref{fig:network}) takes in both $\uparrow \hspace{-0.1cm} \mathbf{I}_{k-1}$ and the warped high-frequency prior $\mathbf{E}^{W_k}$ (warped according to the estimated dense correspondence field).
All the inputs ($\uparrow \hspace{-0.1cm} \mathbf{I}_{k-1}$ and $\mathbf{E}^{W_k}$) as well as $\mathbf{G}_A$ and $\mathbf{G}_B$ are fed into the gate sub-network (cyan in Fig.~\ref{fig:network}) for predicting $\mathbf{G}_\lambda$ and the final high-res output $\mathbf{G}$.

We now introduce the high-frequency prior and the training procedure of the proposed gated bi-network.

\vspace{0.15cm}
\noindent \textbf{High-frequency prior}.
We define high-frequency prior as the indication for location with high-frequency details.
In this work, we generate high-frequency prior maps to enforce spatial guidance for hallucination.
The prior maps are obtained from the mean face template domain.
More precisely, for each training image, we compute the residual image between the original image $\hat{\mathbf{I}}$ and the bicubic interpolation of $\mathbf{I}_{0}$, and then warp the residual map into the mean face template domain.
We average the magnitude of the warped residual maps over all training images and form the preliminary high-frequency map.
To suppress the noise and provide a semantically meaningful prior, we cluster the preliminary high-frequency map into $C$ continuous contours (10 in our implementation).
We form a $C$-channel maps, with each channel carrying one contour.
We refer this $C$-channel maps as our high-frequency prior, and denote it as $E_k(\mathbf{z}): M_k \rightarrow \mathbb{R}^C$.
We use $\mathbf{E}_k$ to represent $E_k(\mathbf{z})$ for all $\mathbf{z} \in M_k$.
An illustration of the prior is shown in Fig.~\ref{fig:prior}(c).

\vspace{0.15cm}
\noindent \textbf{Learning the gated bi-network}.
We train the three parts of convolutional neural networks to predict $\mathbf{G}_A$, $\mathbf{G}_B$ and $\mathbf{G}_\lambda$ in our unified bi-network architecture.
Each part of the network has a distinct training loss.
For training the common branch, we use the following loss over all training samples
\begin{align}
L_A = \|\hat{\mathbf{I}}_k - \uparrow \hspace{-0.1cm} \mathbf{I}_{k-1} - \mathbf{G}_A\|_F^2.
\label{eqn:lossa}
\end{align}
The high-frequency branch has two inputs: $\uparrow \hspace{-0.1cm} \mathbf{I}_{k-1}$ and the \textit{warped} high-frequency prior $\mathbf{E}^{W_k}$ (see Fig.~\ref{fig:prior}(d) for illustration) to predict the output $\mathbf{G}_B$. The two inputs are fused in the channel dimension to form a $(1+C)$-channel input. We use the following loss over all training samples
\begin{align}
L_B = \sum_{c=1}^C \|(\mathbf{E}^{W_k})_c \otimes (\hat{\mathbf{I}}_k - \uparrow \hspace{-0.1cm} \mathbf{I}_{k-1} - \mathbf{G}_B)\|_F^2,
\label{eqn:lossb}
\end{align}
where $(\mathbf{E}^{W_k})_c$ denotes the $c$-th channel of the warped high-frequency prior maps.
Compared to the common branch, we additionally utilize the prior knowledge as input and only penalize over the high-frequency area.
Learning to predict the gate map $\mathbf{G}_\lambda$ is supervised by the final loss
\begin{align}
L = \|\hat{\mathbf{I}}_k - \uparrow \hspace{-0.1cm} \mathbf{I}_{k-1} - \mathbf{G}\|_F^2.
\label{eqn:lossall}
\end{align}
%

We train the proposed gated bi-network with three steps. Step~\textit{i}: We only enable the supervision from $L_A$ (Eq.~\ref{eqn:lossa}) to pre-train the common branch; Step~\textit{ii}: We only enable $L_B$ (Eq.~\ref{eqn:lossb}) to pre-train the high-frequency branch; Step~\textit{iii}: We finally fine-tune the whole gated bi-network with the supervision from $L$ (Eq.~\ref{eqn:lossall}). In the last step, we set the learning rate of the parameters related to the gate map to be 10 times as the parameters in the two branches. Note that we can still use back-propagation to learn the whole bi-network in our last step.

\subsection{$f_k$ - Dense field deformation coefficients prediction}
\label{flow}

We apply a simple yet effective strategy to update the correspondence field coefficients estimation ($f_k$).
Observing that predicting a sparse set of facial landmarks is more robust and accurate under low resolution, we transfer the facial landmarks deformation coefficients to the dense correspondence field.
More precisely, we simultaneously obtain two sets of $N$ deformation bases: $\mathbf{B}_k(\mathbf{z}) \in \mathbb{R}^{2 \times N}$ for the dense field, and $\mathbf{S}_k(l) \in \mathbb{R}^{2 \times N}$ for the landmarks, where $l$ is the landmark index.
The bases for the dense field and landmarks are one-to-one related, i.e. both $\mathbf{B}_k(\mathbf{z})$ and $\mathbf{S}_k(l)$ share the same deformation coefficients $\mathbf{p}_k \in \mathbb{R}^N$:
\begin{equation}
W_k(\mathbf{z}) = \mathbf{z} + \mathbf{B}_k(\mathbf{z}) \mathbf{p}_k; ~ \mathbf{x}_k(l) = \bar{\mathbf{x}}_k(l) + \mathbf{S}_k(l) \mathbf{p}_k,
\end{equation}
where $\mathbf{x}_k(l) \in \mathbb{R}^2$ denotes the coordinates of the $l$-th landmark, and $\bar{\mathbf{x}}_k(l)$ denotes its mean location.

To predict the deformation coefficients $\mathbf{p}_k$ in each cascade $k$, we utilize the powerful cascaded regression approach~\cite{tzimiropoulos2015project} for estimation.
A Gauss-Newton steepest descent regression matrix $\mathbf{R}_k$ is learned in each iteration $k$ to map the observed appearance to the deformation coefficients update:
\begin{align}
\mathbf{p}_k = \mathbf{p}_{k-1} + f_k(\mathbf{I}_{k-1}; \mathbf{p}_{k-1}) = \mathbf{p}_{k-1} + \mathbf{R}_k (\phi(\mathbf{I}_{k-1}; \mathbf{x}_{k-1}(l)|_{l=1,...,L}) - \bar{\phi}),
\end{align}
where $\phi$ is the shape-indexed feature~\cite{dollar2010cascaded,cao2014face} that concatenates the local appearance from all $L$ landmarks, and $\bar{\phi}$ is its average over all the training samples.

To learn the Gauss-Newton steepest descent regression matrix $\mathbf{R}_k$, we follow \cite{tzimiropoulos2015project} to learn the Jacobian $\mathbf{J}_k$ and then obtain $\mathbf{R}_k$ via constructing the project-out Hessian: $\mathbf{R}_k = (\mathbf{J}_k^\top \mathbf{J}_k)^{-1} \mathbf{J}_k^\top$. We refer readers to \cite{tzimiropoulos2015project} for more details.

It is worth mentioning that the face flow method~\cite{snape2015face} that applies a landmark-regularized Lucas-Kanade variational minimization~\cite{baker2004lucas} is also a good alternative to our problem. Since we have obtained satisfying results with our previously introduced deformation coefficients transfer strategy, which is purely discriminative and much faster than face flow (8ms per cascade in our approach v.s. 1.4s for face flow), we use the coefficients transfer approach in our experiments.

\subsection{Implementation Details}
\label{imple}

\setlength{\tabcolsep}{1.7pt}
\begin{table}[t]
\centering
\caption{The architecture of the bi-network in the first cascade.}
\label{table:net1}
\scriptsize
\begin{tabular}{|c|c|c|c|c|c|c|c|c|}
\hline
Network                                                                                         & \begin{tabular}[c]{@{}c@{}}Layer Index\\ (Depth)\end{tabular} & \begin{tabular}[c]{@{}c@{}}Kernel\\ Size\end{tabular} & Stride & Pad & \begin{tabular}[c]{@{}c@{}}Output\\ Channels\end{tabular} & Rectifier & \begin{tabular}[c]{@{}c@{}}Learining Rate\\ (Pre-train)\end{tabular} & \begin{tabular}[c]{@{}c@{}}Learning Rate\\ (End-to-end)\end{tabular} \\ \hline \hline
\multirow{4}{*}{\begin{tabular}[c]{@{}c@{}}Common\\ Sub-net\\ (24 layers)\end{tabular}}         & 1-4                                                           & 3 $\times$ 3                                          & 1      & 1   & 64                                                        & ReLU      & $10^{-4}$                                                            & $10^{-5}$                                                            \\
                                                                                                & 5-20                                                          & 3 $\times$ 3                                          & 1      & 1   & 128                                                       & ReLU      & $10^{-4}$                                                            & $10^{-5}$                                                            \\
                                                                                                & 21-23                                                         & 3 $\times$ 3                                          & 1      & 1   & 32                                                        & ReLU      & $10^{-4}$                                                            & $10^{-5}$                                                            \\
                                                                                                & 24                                                            & 3 $\times$ 3                                          & 1      & 1   & 1                                                         & /         & $10^{-5}$                                                            & $10^{-6}$                                                            \\ \hline \hline
\multirow{4}{*}{\begin{tabular}[c]{@{}c@{}}High-frequency\\ Sub-net\\ (24 layers)\end{tabular}} & 1-4                                                           & 3 $\times$ 3                                          & 1      & 1   & 64                                                        & ReLU      & $10^{-4}$                                                            & $10^{-5}$                                                            \\
                                                                                                & 5-20                                                          & 3 $\times$ 3                                          & 1      & 1   & 128                                                       & ReLU      & $10^{-4}$                                                            & $10^{-5}$                                                            \\
                                                                                                & 21-23                                                         & 3 $\times$ 3                                          & 1      & 1   & 32                                                        & ReLU      & $10^{-4}$                                                            & $10^{-5}$                                                            \\
                                                                                                & 24                                                            & 3 $\times$ 3                                          & 1      & 1   & 1                                                         & /         & $10^{-5}$                                                            & $10^{-6}$                                                            \\ \hline \hline
\multirow{2}{*}{\begin{tabular}[c]{@{}c@{}}Gate Network\\ (6 layers)\end{tabular}}              & 1-5                                                           & 3 $\times$ 3                                          & 1      & 1   & 64                                                        & ReLU      & /                                                                    & $10^{-4}$                                                            \\
                                                                                                & 6                                                             & 3 $\times$ 3                                          & 1      & 1   & 1                                                         & /         & /                                                                    & $10^{-5}$                                                            \\ \hline
\end{tabular}
\end{table}
\setlength{\tabcolsep}{1.4pt}

\setlength{\tabcolsep}{1.7pt}
\begin{table}[t]
\centering
\caption{The architecture of the bi-network in the subsequent cascades.}
\label{table:net2}
\scriptsize
\begin{tabular}{|c|c|c|c|c|c|c|c|c|}
\hline
Network                                                                                         & \begin{tabular}[c]{@{}c@{}}Layer Index\\ (Depth)\end{tabular} & \begin{tabular}[c]{@{}c@{}}Kernel\\ Size\end{tabular} & Stride & Pad & \begin{tabular}[c]{@{}c@{}}Output\\ Channels\end{tabular} & Rectifier & \begin{tabular}[c]{@{}c@{}}Learining Rate\\ (Pre-train)\end{tabular} & \begin{tabular}[c]{@{}c@{}}Learning Rate\\ (End-to-end)\end{tabular} \\ \hline \hline
\multirow{4}{*}{\begin{tabular}[c]{@{}c@{}}Common\\ Sub-net\\ (12 layers)\end{tabular}}         & 1-4                                                           & 3 $\times$ 3                                          & 1      & 1   & 64                                                        & ReLU      & $10^{-5}$                                                            & $10^{-6}$                                                            \\
                                                                                                & 5-8                                                           & 3 $\times$ 3                                          & 1      & 1   & 128                                                       & ReLU      & $10^{-5}$                                                            & $10^{-6}$                                                            \\
                                                                                                & 9-11                                                          & 3 $\times$ 3                                          & 1      & 1   & 32                                                        & ReLU      & $10^{-5}$                                                            & $10^{-6}$                                                            \\
                                                                                                & 12                                                            & 3 $\times$ 3                                          & 1      & 1   & 1                                                         & /         & $10^{-6}$                                                            & $10^{-7}$                                                            \\ \hline \hline
\multirow{4}{*}{\begin{tabular}[c]{@{}c@{}}High-frequency\\ Sub-net\\ (12 layers)\end{tabular}} & 1-4                                                           & 3 $\times$ 3                                          & 1      & 1   & 64                                                        & ReLU      & $10^{-5}$                                                            & $10^{-6}$                                                            \\
                                                                                                & 5-8                                                           & 3 $\times$ 3                                          & 1      & 1   & 128                                                       & ReLU      & $10^{-5}$                                                            & $10^{-6}$                                                            \\
                                                                                                & 9-11                                                          & 3 $\times$ 3                                          & 1      & 1   & 32                                                        & ReLU      & $10^{-5}$                                                            & $10^{-6}$                                                            \\
                                                                                                & 12                                                            & 3 $\times$ 3                                          & 1      & 1   & 1                                                         & /         & $10^{-6}$                                                            & $10^{-7}$                                                            \\ \hline \hline
\multirow{2}{*}{\begin{tabular}[c]{@{}c@{}}Gate Network\\ (6 layers)\end{tabular}}              & 1-5                                                           & 3 $\times$ 3                                          & 1      & 1   & 64                                                        & ReLU      & /                                                                    & $10^{-5}$                                                            \\
                                                                                                & 6                                                             & 3 $\times$ 3                                          & 1      & 1   & 1                                                         & /         & /                                                                    & $10^{-6}$                                                            \\ \hline
\end{tabular}
\end{table}
\setlength{\tabcolsep}{1.4pt}

We train a deep convolutional bi-network for each cascade.
The bi-network is divided into three sub-nets: common sub-net, high-frequency sub-net and the gate sub-net.
Due to the flexibility of the proposed cascaded framework, the depth for each cascade can be different.
As a trade-off between accuracy and speed, we employ a deeper structure (24 layers for the common and high-frequency sub-nets) in the first cascade, while 12 layers for the subsequent cascades.
Please refer to Tab.~\ref{table:net1} (first cascade) and Tab.~\ref{table:net2} (subsequent cascades) for the detailed network structure.
Note that no pooling layer is used in our architecture.

When predicting the dense correspondence field, the appearance vector for each sample is obtained by concatenating the local SIFT descriptor of all facial key points in order to keep consistent with \cite{tzimiropoulos2015project}.
We do not claim that it is the optimal appearance representation. 
Given the limited amount of training data (only 2811 with facial points annotations), we empirically observed that such representation is powerful enough for the discriminative prediction.
Before the dense field prediction in the first cascade, the face is transformed to a reference frame based on the predicted eyes location (by a common 6-layer CNN).
%
We used eye locations since their predictions are more robust and stable under low-res condition.
For the face hallucination step in each cascade, we add a back-projection regularization to suppress error accumulation among cascades.
The effect brought by such regularization is specifically significant in the subsequent cascades.

\section{Experiments}
\label{exp}
\vspace{-0.27cm}
\noindent \textbf{Datasets}. Following \cite{yang2013structured,jin2015robust}, we choose the following datasets that contain both in-the-wild and lab-constrained faces with various poses and illuminations.

%
\begin{enumerate}
  \item \textit{MultiPIE}~\cite{gross2010multi} was originally proposed for face recognition. A total of more than 750,000 faces from 337 identities are collected under lab-constrained environment. We use the same 351 images as used in \cite{yang2013structured} for evaluation.

  \item \textit{BioID}~\cite{jesorsky2001robust} contains 1521 faces also collected in the constrained settings. We use the same 100 faces as used in \cite{jin2015robust} for evaluation.

  \item \textit{PubFig}~\cite{kumar2009attribute} contains 42461 faces (the evaluation subset) from 140 identities originally for evaluating face verification and later used for evaluating face hallucination~\cite{yang2013structured}. The faces are collected from the web and hence in-the-wild. Due to the existence of invalid URLs, we use a total of 20991 faces for evaluation. Further, following \cite{jin2015robust}, we use \textit{PubFig83}~\cite{pinto2011scaling}, a subset of PubFig with 13838 images, to experiment with input blurred by unknown Gaussian kernel. Similar to \cite{jin2015robust}, we test with the same 100-image-subset of PubFig83.

  \item \textit{Helen}~\cite{le2012interactive} contains 2330 in-the-wild faces with high resolution. The mean face size is as large as 275pxIOD. We evaluate with the 330-image test set.
\end{enumerate}

\noindent \textbf{Metric}. We follow existing studies~\cite{jin2015robust,yang2013structured,liu2007face,dong2015image,wang2015deep} to adopt PSNR (dB) and only evaluate on the luminance channel of the facial region. The definition of the facial region is the same as used in \cite{jin2015robust}. Similar to \cite{jin2015robust}, SSIM is not reported for in-the-wild faces due to irregular facial shape.




\noindent \textbf{Implementation details}. 
Our framework consists of $K=4$ cascades, and each cascade has its specific learned network parameters and Gauss-Newton steepest descent regression matrix. During training, our model requires two parts of training data, one for training the cascaded dense face correspondence field, and the other for training the cascaded gated bi-networks for hallucination. The model is trained by iterating between these two parts of the training data. For the former part, we use the training set from 300W~\cite{sagonas2013300} (the same 2811 images used in \cite{tzimiropoulos2015project}) for estimating deformation coefficient and BU4D~\cite{zhang2014bp4d,zhang2013high} dataset for obtaining dense face correspondence basis (following \cite{snape2015face}). For the latter part, as no manual labeling is required, we leverage the existing large face database CelebA~\cite{liu2015deep} for training the gated bi-network.

\vspace{-0.4cm}
\subsection{Comparison with state-of-the-art methods}
\label{soa}
\vspace{-0.2cm}

We compare our approach with two types of methods: (I) general super resolution (SR) approaches and (II) face hallucination approaches.
%
%
For SR methods, we compare with the recent state-of-the-art approaches~\cite{timofte2014a+,dong2015image,wang2015deep,salvador2015naive} based on the original released codes.
For face hallucination methods, we report the result of \cite{jin2015robust,capel2001super,liu2007face} by directly referring to the literature~\cite{jin2015robust}. We compare with \cite{yang2013structured,ma2010hallucinating} by following the implementation of \cite{yang2013structured}. We re-transform the input face to canonical-view if the method assumes the input must be aligned. Hence, such method would enjoy extra advantages in the comparison. If the method requires exemplars, we feed in the same in-the-wild samples in our training set. We observe that such in-the-wild exemplars improve the exemplar-based baseline methods compared to their original implementation. Codes for \cite{tappen2012bayesian} is not publicly available. Similar to \cite{jin2015robust}, we provide the qualitative comparison with \cite{tappen2012bayesian}.

We conduct the comparison in two folds: 1. The input is the down-sampled version of the original high-res image as many of the previous SR methods are evaluated on~\cite{tappen2012bayesian,timofte2014a+,dong2015image,wang2015deep,salvador2015naive} (referred as the conventional SR setting, Sec. 4.1.1); 2. The input is additionally blurred with unknown Gaussian kernel before down-sampling as in~\cite{jin2015robust,liu2007face,yang2013structured} (referred as the Gaussian-blurred setting, Sec. 4.1.2).

\noindent \textbf{4.1.1~~The conventional SR evaluation setting}.
We experiment with two scenarios based on two different types of input face size configuration:
\vspace{-0.15cm}
\begin{enumerate}
  \item \textit{Fixed up-scaling factors} -- The input image is generated by resizing the original image with a fixed factor. For MultiPIE, following \cite{yang2013structured} we choose the fixed factor to be 4. For the in-the-wild datasets (PubFig and Helen), we evaluate for scaling factors of 2, 3 and 4 as in \cite{wang2015deep,dong2015image,salvador2015naive,timofte2014a+} (denoted as $2\times, 3\times, 4\times$ respectively in Tab.~\ref{table:soa1}). In this case, different inputs might have different face sizes. The proposed CBN is flexible to handle such scenario. Other existing face hallucination approaches \cite{capel2001super,liu2007face,ma2010hallucinating,yang2013structured} cannot handle different input face sizes and their results in this scenario are omitted.
  \item \textit{Fixed input face sizes} -- Similar to the face hallucination setting, the input image is generated by resizing the original image to ensure the input face size to be fixed (e.g. 5 or 8 pxIOD, denoted as 5/8px in Tab.~\ref{table:soa1}). Hence, the required up-scaling factor is different for each input. For baseline approaches, \cite{wang2015deep} can naturally handle any up-scaling requirement. For other approaches, we train a set of models for different up-scaling factors. During testing, we pick up the most suitable model based on the specified up-scaling factor.
\end{enumerate}
\vspace{-0.15cm}

We need to point out that the latter scenario is more challenging and appropriate for evaluating a face hallucination algorithm, because recovering the details of the face with the size of 5/8pxIOD is more applicable for low-res face processing applications. In the former scenario, the input face is not small enough (as revealed in the bicubic PSNR in Tab.~\ref{table:soa1}), such that it is more like a facial image enhancement problem rather than the challenging face hallucination~task.

\setlength{\tabcolsep}{4pt}
\begin{table}[t]
\begin{center}
\caption{Results under the conventional SR setting (for Sec~4.1.1). Numbers in the parentheses indicate SSIM and the remaining represent PSNR (dB). The first part of the results are from Scenario 1 where each method super-resolves for a fixed factor (2$\times$, 3$\times$ or 4$\times$), while the latter part are from Scenario 2 that each method begins from the same face size (5 or 8 pxIOD, i.e. the inter-ocular distance is 5 or 8 pixels). The omitted results (-) are due to their incapability of handling varying input face size.}
\label{table:soa1}
\scriptsize
\vspace{-0.3cm}
\begin{tabular}{|c|c|c|c|c|c|c|c|c|c|c|}
\hline
\multirow{3}{*}{Dataset}  & \multirow{3}{*}{\begin{tabular}[c]{@{}c@{}}Input\\ Size\end{tabular}} & \multirow{3}{*}{Bicubic} & \multicolumn{4}{c|}{(I) General super-resolution} & \multicolumn{3}{c|}{(II) Face hallucination}     & \multirow{3}{*}{\textbf{CBN}} \\ \cline{4-10}
                          &                                                                       &                          & A+         & SRCNN      & CSCN       & NBF        & PCA    & \multirow{2}{*}{\cite{ma2010hallucinating}} & \multirow{2}{*}{\cite{yang2013structured}} &                                                                         \\
                          &                                                                       &                          & \cite{timofte2014a+}          & \cite{dong2015image}          & \cite{wang2015deep}          & \cite{salvador2015naive}          & \cite{capel2001super,liu2007face}      &                    &                    &                                                                         \\ \hline \hline
\multirow{2}{*}{MultiPIE} & \multirow{2}{*}{4$\times$}                                                    & 33.66                    & 34.53      & 34.75      & 35.10      & 34.73      & 33.98       & 34.07                   & 34.31                   & \textbf{35.65}                                                                        \\
                          &                                                                       & (.900)                   & (.910)     & (.913)     & (.920)     & (.912)     & (.904) & (.907)             & (.903)             & (\textbf{.926})                                                                  \\ \hline
\multirow{3}{*}{PubFig}   & 2$\times$                                                                     & 34.78                    & 35.89      & 36.12      & 36.47      & 35.98      & -      & -                  & -                  & \textbf{36.66}                                                                        \\
                          & 3$\times$                                                                     & 31.52                    & 32.02      & 32.13      & 32.88      & 32.09      & -      & -                  & -                  & \textbf{33.17}                                                                        \\
                          & 4$\times$                                                                     & 29.61                    & 30.02      & 30.15      & 30.79      & 30.16      & -      & -                  & -                  & \textbf{31.28}                                                                        \\ \hline
\multirow{3}{*}{HELEN}    & 2$\times$                                                                     & 41.96                    & 42.77      & 42.95      & 43.37      & 43.01      & -      & -                  & -                  & \textbf{43.51}                                                                        \\
                          & 3$\times$                                                                     & 38.52                    & 38.89      & 39.10      & 39.57      & 39.15      & -      & -                  & -                  & \textbf{39.78}                                                                        \\
                          & 4$\times$                                                                     & 36.59                    & 36.81      & 36.87      & 37.61      & 36.89      & -      & -                  & -                  & \textbf{37.94}                                                                        \\ \hline \hline
\multirow{2}{*}{MultiPIE} & \multirow{2}{*}{5px}                                                  & 25.39                    & 25.63      & 25.72      & 25.93      & 25.75      & 25.62       & 25.83                   &  25.72                  & \textbf{27.14}                                                                   \\
                          &                                                                       & (.752)                   & (.767)     & (.771)     & (.773)     & (.769)     & (.767)       & (.774)                   & (.769)                   & (\textbf{.808})                                                                  \\ \hline
\multirow{2}{*}{PubFig}                    & 8px  & 22.32 & 22.79 & 22.98 & 23.25 & 23.08 & 23.37 & 23.57 & 23.10 & \textbf{26.83}                                                                   \\ & 5px                                                                   & 20.63                    & 20.96      & 21.07      & 21.33      & 21.04      & 21.42       & 21.58                   & 21.19                   & \textbf{25.31}                                                                   \\ \hline
\multirow{2}{*}{HELEN}                     & 8px & 21.86 & 22.24 & 22.47 & 22.69 & 22.53 & 22.95 & 23.01 & 22.62 & \textbf{26.36}                                                                   \\ & 5px                                                                   & 20.28                    & 20.50      & 20.59      & 20.84      & 20.57      & 21.09       & 21.13                   & 20.64                   & \textbf{25.09}                                                                   \\ \hline
\end{tabular}
\vspace{-0.6cm}
\end{center}
\end{table}
\setlength{\tabcolsep}{1.4pt}
\setlength{\tabcolsep}{4pt}
\begin{table}[t]
\begin{center}
\caption{Results under the Gaussian-blur setting (for Sec. 4.1.2). Numbers in parentheses indicate SSIM and the remaining represent PSNR (dB). Settings adhere to \cite{jin2015robust}. For a fair comparison, we feed in the same number of \textit{in-the-wild} exemplars from CelebA when evaluating \cite{yang2013structured}, instead of the originally used MultiPIE in the released codes.}
\label{table:soa2}
\scriptsize
\vspace{-0.3cm}
\begin{tabular}{|c|c|c|c|c|c|c|c|c|c|c|}
\hline
\multirow{3}{*}{Dataset} & \multirow{3}{*}{Bicubic} & \multicolumn{4}{c|}{(I) General super-resolution} & \multicolumn{4}{c|}{(II) Face hallucination}                          & \multirow{3}{*}{\textbf{CBN}} \\ \cline{3-10}
                         &                          & A+         & SRCNN      & CSCN       & NBF        & PCA    & \multirow{2}{*}{\cite{ma2010hallucinating}} & \multirow{2}{*}{\cite{yang2013structured}} & \multirow{2}{*}{\cite{jin2015robust}} &                      \\
                         &                          & \cite{timofte2014a+}          & \cite{dong2015image}          & \cite{wang2015deep}          & \cite{salvador2015naive}          & \cite{capel2001super,liu2007face}      &                    &                    &                    &                      \\ \hline
\multirow{2}{*}{BioID}   & 19.67                    & 20.47           & 20.59           & 20.86           & 20.60           & 21.51  & 21.77                   & 20.01                   & 22.32              & \textbf{24.55}                     \\
                         & (.670)                   & (.684)     & (.685)     & (.695)     & (.688)     & (.770) & (.776)             & (.689)             & (.810)             & (\textbf{.852})               \\ \hline
PubFig83                 & 24.78                    & 25.20           & 25.22      & 25.65      & 25.47           & 25.72       & 25.83                   & 25.02              & 26.17              & \textbf{29.83}                \\ \hline
\end{tabular}
\vspace{-0.6cm}
\end{center}
\end{table}
\setlength{\tabcolsep}{1.4pt}
\setlength{\tabcolsep}{4pt}
\begin{table}[!h]
\begin{center}
\caption{PSNR results (dB) of in-house comparison of the proposed CBN (for Sec.~\ref{inhouse}).}
\scriptsize
\vspace{-0.3cm}
\label{table:inhouse}
\begin{tabular}{|c|c|c|c|c|c|}
\hline
Dataset  & \begin{tabular}[c]{@{}c@{}}1a. Only Common Branch\\ i.e. Vanilla Cascaded CNN\end{tabular} & \begin{tabular}[c]{@{}c@{}}1b. Only High-\\ Freq. Branch\end{tabular} & \begin{tabular}[c]{@{}c@{}}2. Fixed\\ Correspondence\end{tabular} & \begin{tabular}[c]{@{}c@{}}3. Single\\ Cascade\end{tabular} & \textbf{\begin{tabular}[c]{@{}c@{}}Full\\ Model\end{tabular}} \\ \hline
PubFig   & 23.76                                                                                      & 24.66                                                                 & 23.85                                                             & 22.09                                                       & \textbf{25.31}                                                \\ \hline
HELEN    & 23.57                                                                                      & 24.53                                                                 & 23.77                                                             & 21.83                                                       & \textbf{25.09}                                                \\ \hline
PubFig83 & 28.06                                                                                      & 29.31                                                                 & 28.34                                                             & 26.70                                                       & \textbf{29.83}                                                \\ \hline
\end{tabular}
\vspace{-0.6cm}
\end{center}
\end{table}
\setlength{\tabcolsep}{1.4pt}
\setlength{\tabcolsep}{4pt}
\begin{table}[!h]
\centering
\caption{PSNR results (dB) of varying input face size: 3, 5, 8, 10 pxIOD (for Sec.~\ref{lowerbound}).}
\label{table:lowerbound}
\scriptsize
\vspace{-0.3cm}
\begin{tabular}{|c|c|c|c|c|c|c|c|c|}
\hline
\multirow{2}{*}{Dataset} & \multicolumn{2}{c|}{3xpIOD} & \multicolumn{2}{c|}{5pxIOD} & \multicolumn{2}{c|}{8pxIOD} & \multicolumn{2}{c|}{10pxIOD} \\ \cline{2-9}
                         & ~Bicubic~        & ~~CBN~~        & ~Bicubic~       & ~~CBN~~         & ~Bicubic~        & ~~CBN~~        & ~Bicubic~         & ~~CBN~~        \\ \hline
PubFig                   & 18.10               & 20.01           & 20.63         & 25.31       & 22.32               & 26.83           & 23.69                & 27.92           \\ \hline
HELEN                    & 17.82               & 19.78           & 20.28         & 25.09       & 21.86               & 26.36           & 23.29                & 27.46           \\ \hline
\end{tabular}
\vspace{-0.6cm}
\end{table}

\begin{figure}[!h]
\centering
\centering
\scriptsize
Bicubic~~~~~~~\cite{liu2007face}~~~~~~~~CSCN~~~~~~~~Bicubic~~~~~~~\cite{liu2007face}~~~~~~~~CSCN~~~~~~~~Bicubic~~~~~~~\cite{liu2007face}~~~~~~~~CSCN
\includegraphics[width=1\linewidth]{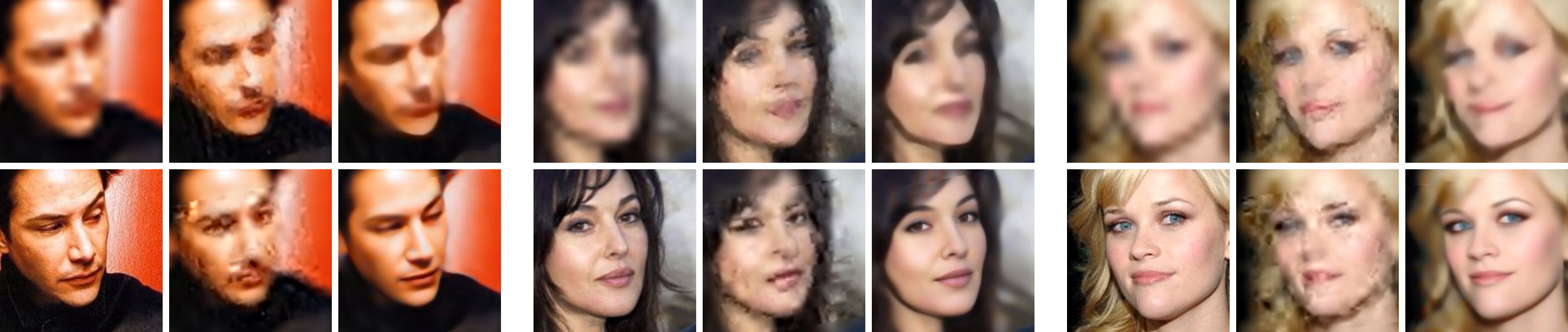}
Original~~~~~~~\cite{yang2013structured}~~~~~~~~~\textbf{CBN}~~~~~~~Original~~~~~~~\cite{yang2013structured}~~~~~~~~\textbf{CBN}~~~~~~~~Original~~~~~~~\cite{yang2013structured}~~~~~~~~\textbf{CBN}~
\vspace{-0.35cm}
\caption{Qualitative results from PubFig/HELEN with input size 5pxIOD (for Sec.~4.1.1, detailed results refer Tab.~\ref{table:soa1}). Best viewed by zooming in the electronic version.}
\label{fig:soa1}
\vspace{0.2cm}
\includegraphics[height=0.6\linewidth]{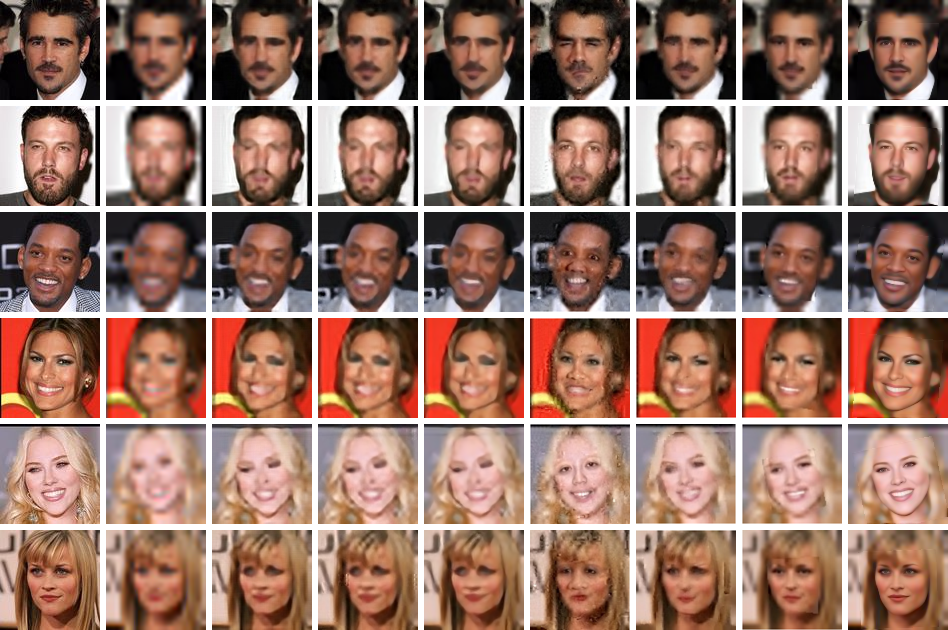}
\scriptsize
Original~~~Bicubic~~~SRCNN~~~~~NBF~~~~~~CSCN~~~~~~~\cite{yang2013structured}~~~~~~~~~~\cite{tappen2012bayesian}~~~~~~~~~~\cite{jin2015robust}~~~~~~~~\textbf{CBN}
\vspace{-0.35cm}
\caption{Qualitative results from the PubFig83 dataset (for Sec.~4.1.2, detailed results refer Tab.~\ref{table:soa2}). The six test samples presented are chosen by strictly following \cite{jin2015robust}.}
\label{fig:soa2}
\vspace{0.2cm}
\includegraphics[width=0.9\linewidth]{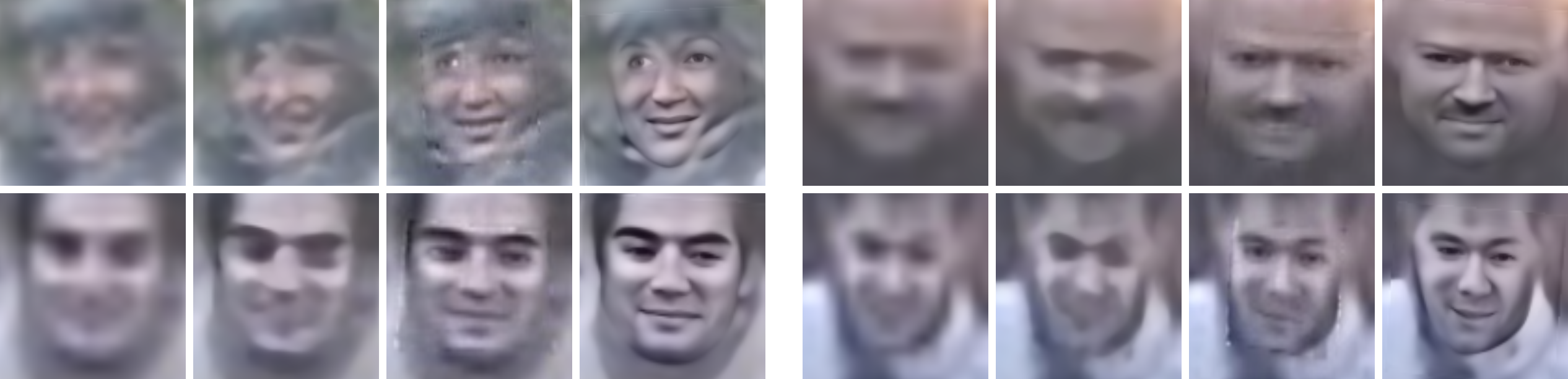}
\scriptsize
~Bicubic~~~~~~CSCN~~~~~~~~\cite{jin2015robust}~~~~~~~~~\textbf{CBN}~~~~~~~~Bicubic~~~~~~CSCN~~~~~~~~~\cite{jin2015robust}~~~~~~~~~\textbf{CBN}
\vspace{-0.35cm}
\caption{Qualitative results for real surveillance videos (for Sec.~4.1.2). The test samples are directly imported from \cite{jin2015robust}. Best viewed by zooming in the electronic version.}
\label{fig:mppca}
\vspace{-0.3cm}
\end{figure}

We report the results in Tab.~\ref{table:soa1}, and provide qualitative results in Fig.~\ref{fig:soa1}. As can be seen from the results, our proposed CBN outperforms all general SR and face hallucination methods in both scenarios. The improvement is especially significant in the latter scenario because our incorporated face prior is more~critical when hallucinating face from very low resolution.
We observe that the general SR algorithms did not obtain satisfying results because they take full efforts to recover only the detectable high-frequency details, which obviously contain noise. In contrast, our approach recovers the details according to the high-frequency prior as well as the estimated dense correspondence field, thus achieving better performance. The existing face hallucination approaches did not perform well either. 
In comparison to the evaluation under the constrained or canonical-view condition (e.g. \cite{yang2013structured}), we found that these algorithms are more likely to fail under in-the-wild setting with substantial shape deformation and appearance variation.
%

\noindent \textbf{4.1.2~~The Gaussian-blur evaluation setting}.
It is also important to explore the capability of handling blurred input images~\cite{efrat2013accurate}.
Our method demonstrates certain degrees of robustness toward unknown Gaussian blur.
%
%
Specifically, in this section, we still adopt the same model as in Sec.~4.1.1, with no extra efforts spent in the training to specifically cope with blurring.
To compare with~\cite{jin2015robust}, we add Gaussian blur to the input facial image in the same way as~\cite{jin2015robust}.
The experimental settings are precisely the same as in \cite{jin2015robust} -
the input faces have the same size (around 8pxIOD); the up-scaling factor is set to be 4;
and $\sigma$ for Gaussian blur kernel is set to be 1.6 for \textit{PubFig83} and 2.4 for \textit{BioID}.
Additional Gaussian noise with $\eta=2$ is added in BioID.
We note that our approach only uses single frame for inference, unlike multiple frames in \cite{jin2015robust}.

We summarize the results in Tab.~\ref{table:soa2}. Qualitative results are shown in Fig.~\ref{fig:soa2}.
From the results it is observed that again CBN significantly outperforms all the compared approaches.
We attribute the robustness toward the unknown Gaussian blur on the spatial guidance provided by the face high-frequency prior.

%

Taking advantages of such robustness of our approach, we further test the proposed algorithm over the faces from real surveillance videos.
In Fig.~\ref{fig:mppca}, we compare our result with \cite{wang2015deep,jin2015robust}.
Note that the presented test cases are directly imported from \cite{jin2015robust}.
Again, our result demonstrates the most appealing visual quality compared to existing state-of-the-art approaches, suggesting the potential of our proposed framework in real-world applications.

\noindent \textbf{4.1.3~~Run time}.
%
%
The major time cost of our approach is consumed on the forwarding process of the gated deep bi-networks.
%
On a single core i7-4790 CPU, the face hallucination steps for the four cascades (from 5pxIOD to 80pxIOD) require 0.13s, 0.17s, 0.70s, 2.76s, respectively.
The time cost of the dense field prediction steps is negligible compared to the hallucination step.
Our framework totally consumes 3.84s, which is significantly faster than existing face hallucination approaches, for examples, 15-20min for \cite{jin2015robust}, 1min for \cite{yang2013structured}, 8min for \cite{liu2007face}, thanks to CBN's purely discriminative inference procedure and the non-exemplar and parametric model structure.

\vspace{-0.3cm}
\subsection{An ablation study}
\label{inhouse}
\vspace{-0.2cm}

%
We investigate the effects of three important components in our framework:
%
\begin{enumerate}
  \item \textbf{Effects of the gated bi-network} (a) We explore the results if we replace the cascaded gated bi-network with the vanilla cascaded CNN, in which only the common branch (the blue branch in Fig.~\ref{fig:network}) is remained. In this case, the spatial information, i.e. the dense face correspondence field is not considered or optimized at all. (b) We also explore the case where only the high-frequency branch (the red branch in Fig.~\ref{fig:network}) is remained.
  \item \textbf{Effects of the progressively updated dense correspondence field} In our framework, the pixel-level correspondence field is refined progressively to better facilitate the subsequent hallucination process. We explore the results if we only use the correspondence estimated from the input low-res image\footnote{As the correspondence estimation is by itself a cascaded process, in this case, we re-order the face corresponding cascades before the super resolution cascades.}. In this case, the spatial configuration estimation is not updated with the growth of the resolution.
  \item \textbf{Effects of the cascade} The cascaded alternating framework is the core for our framework. We explore the results if we train one network and directly super resolve the input to the required size. High-frequency prior is still used in this baseline. We observe an even worse result without this prior.
\end{enumerate}
%
We present the results in Tab.~\ref{table:inhouse}. The experimental setting follows the same setting in Sec.~\ref{soa} - The PubFig and HELEN datasets super-resolve from 5pxIOD while the PubFig83 dataset up-scales 4 times with unknown Gaussian blur.
The results suggest that all components are important to our proposed approach.
%

\vspace{-0.3cm}
\subsection{Discussion}
\label{failure}
\vspace{-0.2cm}

\begin{figure}[t]
\centering
\includegraphics[width=\linewidth]{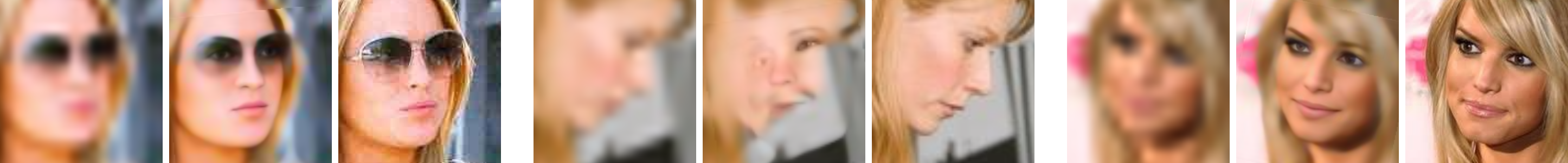}
\scriptsize
\medskip
(a)Bicubic~(a)\textbf{CBN}~(a)Original~~~(b)Bicubic~(b)\textbf{CBN}~(b)Original~~~(c)Bicubic~(c)\textbf{CBN}~(c)Original
\medskip
\vspace{-0.7cm}
\caption{Three types of representative failure cases of our approach (for Sec.~\ref{failure}).}
\label{fig:failureshowoff}
\vspace{0.2cm}
\includegraphics[width=\linewidth]{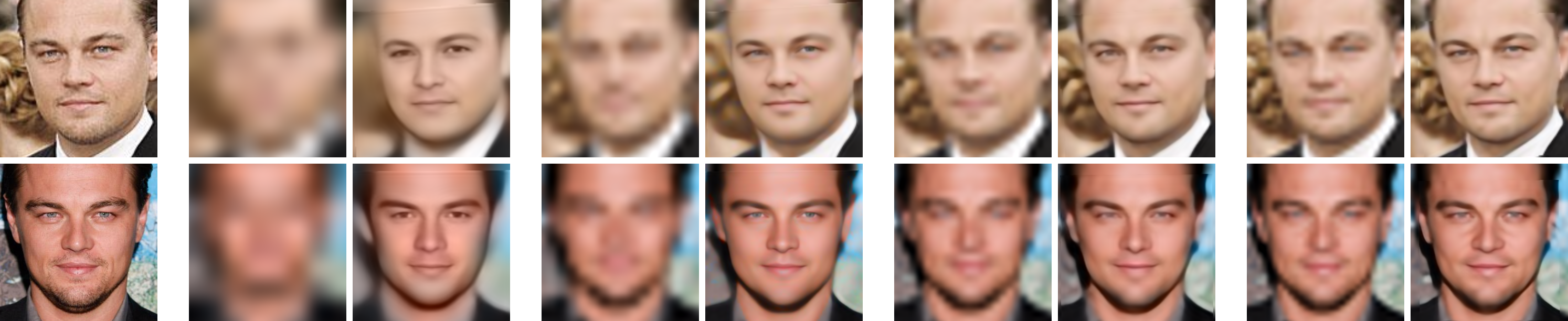}
\medskip
Original~~~~~~~~~~~~3pxIOD~~~~~~~~~~~~~~~~~~~5pxIOD~~~~~~~~~~~~~~~~~~~8pxIOD~~~~~~~~~~~~~~~~~~10pxIOD~~~~~
\medskip
\vspace{-0.7cm}
\caption{Visualization of the result of various input resolution: 3, 5, 8, 10 pxIOD (for Sec.~\ref{lowerbound}). For each resolution, the left is the bicubic result while the right is the output from the proposed CBN framework.}
\label{fig:lowerbound}
\vspace{-0.3cm}
\end{figure}

Despite the effectiveness of our method, we still observe a small set of failure cases.
%
%
Figure~\ref{fig:failureshowoff} illustrates three typical types of failure:
(1) Over-synthesis of occluded facial parts, e.g., the eyes in Fig.~\ref{fig:failureshowoff}(a). In this case, the gate network might have been misled by the light-colored sun-glasses and therefore favours the results from the high-frequency branch.
(2) Ghosting effect, which is caused by inaccurate spatial prediction under low-res. It is rather challenging to localize facial parts with very large head pose in the low-res image.
(3) Incorrect details such as gaze direction. We found that there is almost no reliable gaze direction information presented in the input. Our method only synthesizes the eyes with the most probable gaze direction.
We leave it as future works to address the aforementioned drawbacks.



\vspace{-0.1cm}
\section{Input resolution lower bound}
\label{lowerbound}
\vspace{-0.05cm}

It is of research interest to explore how small an input face we can recover for the in-the-wild settings.
%
%
To achieve this goal, we train a set of models that hallucinates the face from varying input sizes.
We select four representative face sizes: 3, 5, 8, 10 pxIOD as the input and observe their corresponding results.
%
%
%

According to the PSNR results in Tab.~\ref{table:lowerbound} and the qualitative results in Fig.~\ref{fig:lowerbound}, we observe that results originated from 3pxIOD are mostly unrealistic and visually dissimilar to the full resolution image. The dense correspondence field is likely to be incorrectly predicted, and very few information is provided in the low-res input.
On the other hand, if the input face size is no smaller than 5pxIOD, the PSNR would enjoy a significant increase (Tab.~\ref{lowerbound}).
Based on such observations, we believe that the input size of 3pxIOD might be below the lower bound where it is very difficult to recover faces from such resolution.
This constitutes the reason why we choose 5/8pxIOD faces as low-res input in Sec.~\ref{exp}.

\vspace{-0.1cm}
\section{Conclusion}
\vspace{-0.05cm}

We have presented a novel framework for hallucinating faces under substantial shape deformation and appearance variation.
Owing to the specific capability to adaptively refine the dense correspondence field and hallucinate faces in an alternating manner, we obtain state-of-the-art performance and visually appealing qualitative results.
Guided by the high-frequency prior, our framework can leverage spatial cues in the hallucination process.


\bibliographystyle{splncs}
\bibliography{short,ctfs_bib}
\end{document}